\documentclass[a4paper,twoside]{article}

\usepackage{subcaption}
\usepackage{calc}
\usepackage{amssymb}
\usepackage{amstext}
\usepackage{amsmath}
\usepackage{amsthm}
\usepackage{multicol}
\usepackage{pslatex}
\usepackage{apalike}
\usepackage{algorithm2e}
\usepackage[bottom]{footmisc}
\usepackage{dsfont}
\usepackage{SCITEPRESS}

\usepackage{booktabs}
\usepackage{siunitx}
\usepackage{graphicx}
\usepackage{url}
\usepackage{xspace}

\newcommand{\ordinalbench}{\textsc{OrdinalBench}\xspace}
\newcommand{\ordinalbenchbf}{\textbf{\textsc{OrdinalBench}}\xspace}

\begin{document}

\title{
  \ordinalbench: A Benchmark Dataset for Diagnosing Generalization Limits in Ordinal Number Understanding of Vision-Language Models
}

\author{\authorname{
    Yusuke Tozaki\sup{1}\orcidAuthor{0009-0005-1189-0058},
    Hisashi Miyamori\sup{1}\orcidAuthor{0000-0003-3833-9512}
  }
  \affiliation{\sup{1}Division of Frontier Informatics,  Kyoto Sangyo University, Kyoto, Japan}
  \email{\{i2486147, miya\}@cc.kyoto-su.ac.jp}
}


\abstract{
  Vision-Language Models (VLMs) have advanced across multimodal benchmarks but still show clear gaps in ordinal number understanding, i.e., the ability to track relative positions and generalize to large indices.
  We present \ordinalbenchbf, a diagnostic benchmark that standardizes ordinal number understanding as an evaluation task for VLMs.
  The core task is $N$-th object identification, defined by a starting reference and traversal rule.
  Task difficulty is controlled along three axes: (i) ordinal magnitude (from small numbers to extreme cases up to 300), (ii) arrangement complexity (from single loops to maze-like paths), and (iii) object count.
  The benchmark provides 39,000 question--answer pairs, each annotated with a ground-truth reasoning trajectory, balanced across difficulty, supporting controlled yet large-scale testing.
  Beyond answer-only evaluation, our framework requires models to generate structured, stepwise traces of the counting process and supplies an open evaluation toolkit measuring both final accuracy and step-level path consistency.
  Zero-shot evaluations of GPT-5, Gemini 2.5 Flash Lite, Qwen2.5-VL, InternVL3.5, and Molmo reveal sharp degradation under large-ordinal and complex-path conditions, highlighting weak generalization despite strong scores on standard multimodal tasks.
  By framing ordinal number understanding as a core target, \ordinalbench offers a reproducible testbed and actionable diagnostics to drive development of VLMs with stronger sequential reasoning.
  All data and code are available at \url{https://ordinalbench.github.io}
}

\onecolumn \maketitle \normalsize \setcounter{footnote}{0} \vfill

\section{\uppercase{Introduction}}
\label{sec:introduction}

\begin{figure*}[t]
	    \centering
	    \includegraphics[width=1.0\linewidth]{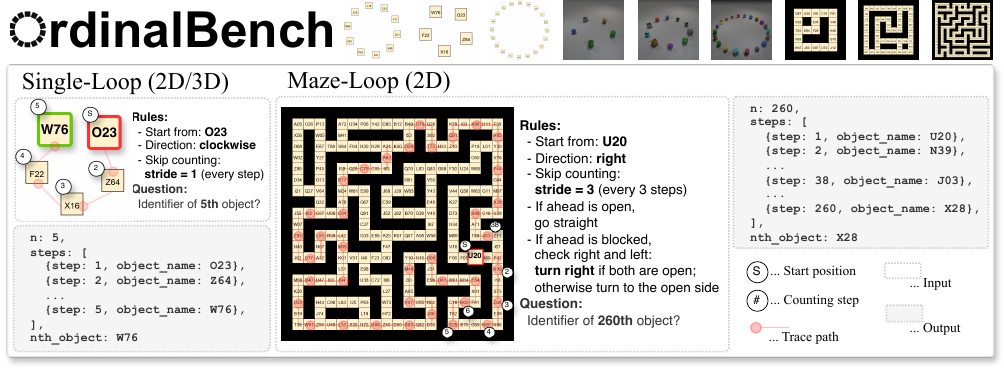}
	    \caption{
			Overview of \ordinalbenchbf.
			We illustrate representative instances of Single-Loop (2D/3D) and Maze-Loop (2D),
			where models follow explicit traversal rules (start position, facing direction, stride, and blocked-turn policy) to count along a path and predict the identifier at the $N$-th position.
			Each example includes the input image and question, as well as a step-wise trace output, enabling evaluation of procedural execution beyond the final answer.
			For brevity, the instruction prompts shown are abridged examples of the actual inputs used in our evaluation.
		}
	    \label{fig:intro}
\end{figure*}

In recent years, Vision-Language Models (VLMs) have advanced markedly and achieve strong performance across multimodal tasks \cite{li2025survey}.
Yet they still exhibit unexpected errors in fundamental visual reasoning, including accurate object enumeration (cardinal understanding) and spatial/ordinal relations \cite{pothiraj2025capture,zhang2025comfort,du2024numclip}.
These failures suggest fragile generalization, potentially amplified by the next-token prediction objective underlying many models \cite{qi2024generalization}.

Generalization weaknesses are particularly salient in \textbf{ordinal number understanding}, which is crucial for real applications yet underexplored.
Ordinals help uniquely identify targets when attributes are insufficient (e.g., ``the black car three vehicles ahead of the white car'') and require sequential tracking, sustained attention, and state updates.
For instance, in user-interface automation, an agent may be instructed to click the $N$-th icon in a toolbar after navigating a menu hierarchy; in embodied robotics, it may need to grasp the $N$-th part along a conveyor by following a belt or wiring route.
However, existing benchmarks are not designed to diagnostically probe the generalization of such procedural ordinal number understanding.

We view ordinal number understanding not as memorized ordinal vocabulary, but as the \textbf{generalization of procedural reasoning}: executing ``find the $N$-th'' step-by-step from visual input.
We therefore introduce \ordinalbench, a diagnostic benchmark that standardizes ordinal number understanding as an evaluation task.
The core task traverses objects from a start point under a rule (e.g., clockwise; a maze traversal rule in Section~\ref{sec:task_definition}) to identify the $N$-th target.
We also introduce skip counting with stride $k>1$ (count every $k$ steps) to further require algorithmic execution.

Nth-object identification comprises two processes: (i) understanding and executing the rule that determines the counting order, and (ii) proceeding correctly until reaching the specified ordinal $N$.
\ordinalbench diagnoses these independently via a spatial axis (simple loops to maze structures) and a numerical axis (small to very large $N$), and additionally controls visual load by varying object count (or grid size).

The contributions of this research are summarized in the following three points.
\begin{enumerate}
    \item We develop and release \ordinalbench, a diagnostic benchmark that systematically evaluates ordinal generalization under rare large $N$, complex loop/maze traversal, and skip counting.
    \item We identify characteristic weaknesses and capability limits of current VLMs via zero-shot evaluation on \ordinalbench.
    \item We discuss implications and future directions for model design, training strategies, and evaluation methods toward more robust VLMs, grounded in observed failure modes.
\end{enumerate}

We describe the benchmark design and evaluation metrics, present experimental results, discuss implications and limitations, and conclude.

\section{\uppercase{Related Work}}
\label{sec:related-work}

Many benchmarks evaluate VLM capabilities, but most do not target the \textbf{generalization of procedural ordinal number understanding} studied here.
Existing benchmarks broadly focus on compositional reasoning or cardinal counting.
For example, generalization benchmarks such as CLEVR \cite{johnson2017clevr} emphasize compositional reasoning over object attributes and relations.
Numerical cognition benchmarks such as TallyQA \cite{acharya2018tallyqa} and HowManyQA \cite{trott2018howmanyqa} focus on static questions of ``how many'' (cardinal understanding).
In contrast, identifying the $N$-th object requires sequential rule application, maintaining internal state, and algorithmic execution, and a framework to systematically evaluate ordinal generalization under large numerical scales and complex paths has been lacking.

\ordinalbench addresses this gap with parameter-controllable synthetic data.
Synthetic data is effective for precise diagnostic evaluation \cite{johnson2017clevr,mumuni2024syntheticdata}, and strict control over factors such as ordinal magnitude and path complexity enables us to identify performance limits.

Prior work also reports that VLMs struggle with number sense and ambiguities in spatial/frame-of-reference reasoning \cite{du2024numclip,pothiraj2025capture,zhang2025comfort}.
\ordinalbench fills the remaining gap of evaluation specifically focused on ordinal generalization and contributes to deeper understanding of VLM capabilities.

\section{\uppercase{Benchmark Dataset}}
\label{sec:benchmark}

\subsection{Design Philosophy}
\label{sec:design_philosophy}

\ordinalbench is a large-scale benchmark designed to diagnose ordinal number understanding in Vision-Language Models (VLMs), particularly the \textbf{generalization limits of procedural reasoning}.
Under controlled synthetic settings, it systematically combines: (i) complex spatial structures such as maze-loops, (ii) large ordinals that rarely appear in training data (e.g., $N\ge 100$), and (iii) algorithmic rules such as skip counting.

\ordinalbench follows the CLEVR-style diagnostic benchmark philosophy \cite{johnson2017clevr,mumuni2024syntheticdata}.
By intentionally removing recognition confounders such as lighting and occlusion, it isolates ordinal number understanding as the evaluation target.
To achieve this goal, we adopt the following design principles.

\paragraph{Targetedness}
We use simple, easily identifiable visual identifiers so that errors can be attributed to reasoning rather than visual ambiguity.

\paragraph{Controllability \& Systematization}
We control difficulty systematically across three axes (arrangement complexity, object count, ordinal magnitude) and additionally introduce skip counting ($k>1$) to disentangle failure factors.

\paragraph{Challenge}
We intentionally include difficult configurations (maze structures, very large ordinals, skip counting) to expose limitations of current SOTA models and clarify directions for improvement.

\paragraph{Scalability}
An automated generation pipeline enables continual creation of harder evaluation sets for future, stronger models \cite{mumuni2024syntheticdata}.

\paragraph{Interpretability}
Beyond final correctness, we design outputs to identify where the reasoning process breaks, enabling detailed analysis of procedural failures.

\subsection{Task Definition}
\label{sec:task_definition}

The core task is \textbf{Nth-object identification}.
Given an image $I$ and traversal rule $R$, a model counts starting from a reference $o_{start}$ (counted as 1st) and identifies the target at ordinal position $N$.
This task is not static pattern recognition; it requires sequential tracking under a rule and maintaining an internal state (the current count), directly probing procedural reasoning.

We formalize the task as a mapping where a VLM $f_{VLM}$ outputs a final target $\hat{o}_N$ and a reasoning trace $\hat{T}$:
\[
(\hat{o}_{N}, \hat{T}) = f_{VLM}(I, R, o_{start}, N, k).
\]
The inputs are:
\begin{itemize}
    \item $I$: input image (e.g., 2D maze, 3D scene)
    \item $R$: traversal rule described in text (e.g., ``go clockwise''; for maze-loops, see the rule below)
    \item $o_{start}$: starting object (counted as 1st)
    \item $N$: target ordinal position ($N\ge 1$)
    \item $k$: counting interval (stride, $k\ge 1$)
\end{itemize}
For maze-loop tasks, we define $R$ as follows: At each step, move forward if the cell ahead is free; otherwise, turn to the preferred side (right/left) if possible, and if that is blocked, turn to the opposite side. Mazes contain no dead ends, so this rule defines a unique next move.
The outputs are:
\begin{itemize}
    \item $\hat{o}_{N}$: predicted $N$-th target
    \item $\hat{T}$: structured reasoning trace $\hat{T}=(\hat{t}_1,\ldots,\hat{t}_N)$
\end{itemize}
In our evaluation, we require models to output $(\hat{o}_N,\hat{T})$ as a structured JSON object to make parsing and scoring unambiguous.

Stride $k$ controls cognitive load.
With $k{=}1$, counting advances at every step.
With $k{>}1$, the model must execute an algorithm that counts ``every $k$ steps.''
Concretely, we count $o_{start}$ as 1st; the position after $k$ steps under $R$ is 2nd, after another $k$ steps is 3rd, and so on.
This diagnoses internal state management and algorithm execution beyond simple instruction following.

By requiring the trace $\hat{T}$, we can quantify where reasoning breaks (nLCP) and how faithful the overall process is (STA) (Section~\ref{sec:evaluation}).

\subsection{Systematic Difficulty Control}
\label{sec:difficulty_control}

\ordinalbench controls difficulty to diagnose generalization limits of procedural reasoning.
The core consists of three independent axes: \textbf{Arrangement Complexity}, \textbf{Object Count}, and \textbf{Ordinal Magnitude}.
This separation helps attribute degradation to spatial structure understanding, working-memory load, or numerical-scale generalization.
We also combine these with skip counting ($k>1$) for more fine-grained diagnostics.

\paragraph{Arrangement Complexity}
We vary the visual and structural complexity of the traversal path to evaluate ordinal tracking under complex visual structures.
\begin{itemize}
    \item \textbf{Level 1 - Single-Loop}: a single closed loop; evaluates basic path tracking.
    \item \textbf{Level 2 - Maze-Loop}: an algorithmically generated complex path (always a loop, with no dead ends); evaluates maintaining ordinal state over hard-to-verbalize global topology.
\end{itemize}

\paragraph{Object Count}
We vary the total number of objects (or grid size) to evaluate attention and working-memory load in larger scenes.
\begin{itemize}
    \item \textbf{Few / Medium / Many}: Single-Loop uses 5/10/20 objects; Maze-Loop uses $7{\times}7/11{\times}11/21{\times}21$ grids.
\end{itemize}

\paragraph{Ordinal Magnitude}
We vary the target ordinal number $N$ to evaluate generalization across numerical scales.
We define a baseline $S$ as ``object count'' for Single-Loop and ``grid size'' for Maze-Loop.
\begin{itemize}
    \item \textbf{Level 1 (Within)}: $2 \le N \le S$
    \item \textbf{Level 2 (Exceed)}: $S < N \le 99$
    \item \textbf{Level 3 (Large Scale)}: $100 \le N \le 300$
\end{itemize}

\subsection{Data Generation Pipeline and Scalability}
\label{sec:data_generation}

\ordinalbench images and QA pairs are automatically generated by parameter-controllable programs.
The pipeline consists of three stages that progressively instantiate the difficulty factors.

\paragraph{Scene Generation}
First, we construct the visual scenes.
2D images are generated with Python libraries, and 3D scenes are built on publicly available codebases (e.g., CLEVR-Ref+ \cite{clevrrefplus_todo}).
To prevent recognition difficulty from becoming noise in reasoning evaluation, we use easily identifiable labels (e.g., \texttt{A00--Z99}) and 48 geometric objects.

\paragraph{Question--Answer Pair Generation}
Second, we generate QA pairs from templates based on the task definition and difficulty axes.
We systematically sample the starting point, traversal rule, target ordinal $N$, and stride $k$ to produce structurally consistent questions.

\paragraph{Reasoning Trace Generation}
Third, for each QA, we simulate the traversal and produce a ground-truth reasoning trace.
The trace records step states (identifier, position, step count, ordinal position, etc.).
Even when $k>1$, we record only the steps where a count occurs (every $k$ steps), so the trace length is always $N$.

\subsection{Dataset Statistics}
\label{sec:dataset_stats}

We generate and release a reproducible evaluation split, \texttt{public\_test}.
It covers 2D Single-Loop, 2D Maze-Loop, and 3D Single-Loop, totaling 2,600 images and 39,000 QA pairs.
The split is balanced across combinations of the difficulty factors (spatial structure, numerical scale, rules, and object count), making it suitable for diagnostic evaluation.

\subsection{Evaluation Protocol and Metrics}
\label{sec:evaluation}

Beyond providing the dataset, \ordinalbench contributes an evaluation protocol and metrics that diagnose the structural correctness of the reasoning process, not only final correctness.
We use the following metrics.

\paragraph{Final Accuracy (Acc@N)}
Final Accuracy measures whether the final prediction $\hat{o}_N$ matches the ground-truth target $o_N$:
\[
\mathrm{Acc@N} = \frac{1}{M}\sum_{i=1}^{M}\mathds{1}\!\left[\hat{o}_{N}^{(i)} = o_{N}^{(i)}\right].
\]

\paragraph{Normalized Longest Correct Prefix (nLCP)}
nLCP measures the normalized length of the longest correct prefix $LCP_i$ of the predicted trace, averaged over samples:
\[
\mathrm{nLCP} = \frac{1}{M}\sum_{i=1}^{M}\frac{LCP_i}{|T_i|}.
\]

\paragraph{Stepwise Trace Accuracy (STA)}
STA evaluates whether each step of the predicted trace matches the ground truth, averaged over steps and samples:
\[
\mathrm{STA} = \frac{1}{M}\sum_{i=1}^{M}\frac{1}{|T_i|}\sum_{t=1}^{|T_i|}\mathds{1}\!\left[\hat{t}_{t}^{(i)} = t_{t}^{(i)}\right].
\]

\paragraph{Trace Coverage (Cov.)}
Trace Coverage measures the fraction of samples for which a model returns at least one valid step in the structured JSON trace:
\[
\mathrm{Cov} = \frac{1}{M}\sum_{i=1}^{M} \mathds{1}\!\left[\,\lvert \hat{T}_i\rvert > 0\,\right].
\]

\paragraph{Evaluation Toolkit}
We provide an evaluation toolkit that takes model outputs (structured JSON traces) and automatically computes Acc@N, nLCP, STA, and Cov., facilitating reproducible comparisons and detailed diagnostic analysis.

\section{\uppercase{Experiments}}
\label{sec:experiments}

\subsection{Experimental Setup}
\label{sec:experimental_setup}

In this section, we describe the experimental setup for evaluating how well current state-of-the-art VLMs can generalize procedural reasoning on \ordinalbench, with a focus on its limitations.
The overall experimental design follows the research objectives presented in the introduction, aiming to fairly and rigorously test generalization to novel scenarios (e.g., complex spatial structures or unfamiliar numbers).

All results reported in this section are obtained on the \texttt{public\_test} split generated and released with this work (Section~\ref{sec:dataset_stats}). For API-based models (GPT-5 and Gemini 2.5), we additionally report results on stratified subsets sampled from \texttt{public\_test} (2,500 samples for each 2D task and 1,500 for 3D) to reduce inference cost. Stratification preserves distributions over ordinal level and stride, and task-specific factors.

\paragraph{Evaluated Models}
We evaluate API-based proprietary models (OpenAI GPT-5 \cite{openai2025gpt5}, Gemini 2.5 Flash Lite \cite{gemini2025v25}) and major, widely used open-source VLM suites (Qwen2.5-VL \cite{bai2025qwen25vl}, InternVL3.5 \cite{internvl35_todo}, Molmo \cite{molmo_todo}).
This selection is justified on the following three points:
(1) to clarify the performance limits at the current state of the art;
(2) to verify whether weaknesses are specific to particular models or are general challenges by comparing models with different architectures and training data;
and (3) to enable future additional evaluations and analyses by including open-source models.

\paragraph{Evaluation Setting}
To evaluate the true generalization capability of the models, all tasks were conducted in a \textbf{zero-shot setting} \cite{qi2024generalization}.
Zero-shot refers to a setting where models tackle tasks at inference time without any prior fine-tuning or in-context examples.
This is a rigorous method for measuring the ability to handle unseen patterns, such as unfamiliar numbers or complex structures.

\paragraph{Implementation Details}
To ensure fair comparison and reproducibility across models, the prompt design and inference parameters were carefully standardized.
Each model was given a basic prompt containing the task rules, traversal method, starting point, and target ordinal number.
This design follows the standard for diagnostic benchmarks, employing an identical format for all models \cite{johnson2017clevr}.
We also instructed the models to output their reasoning in a structured JSON format as described in Section~\ref{sec:task_definition}.
For inference parameters, the temperature was fixed at 0.0 to ensure deterministic outputs.
This allows for an accurate comparison of the models' intrinsic reasoning capabilities, minimizing the influence of prompt design differences and sampling stochasticity.

\paragraph{Evaluation Metrics}
We report the four metrics defined in Section~\ref{sec:evaluation}: Final Accuracy (Acc@N) and three trace-based metrics (nLCP, STA, and Cov.) to assess procedural faithfulness and output validity.
Unless noted, all metrics are macro-averaged across QA instances.

\subsection{Quantitative Results and Analysis}
    In this section, we quantitatively analyze the performance of VLMs evaluated using \ordinalbench.
    First, we provide an overview of the overall performance of all models, and then we conduct a detailed performance analysis along each difficulty axis to reveal the limits of current VLMs' ordinal number understanding capabilities.

    \subsubsection{Overall Performance Overview}

    Table~\ref{tab:main_results} summarizes the overall scores of the four evaluation metrics (Acc@N, nLCP, STA, and Cov.) for all evaluated models across three task categories: 2D Single-Loop, 3D Single-Loop, and 2D Maze-Loop.
    We first observe a \textbf{dramatic impact of task complexity on model performance}.
    In the simplest 2D Single-Loop task, Qwen2.5-VL-32B achieved the highest Acc@N of 32.83\%, and several models significantly surpassed the Chance Level (approx. 11.7\%).
    However, in the 3D Single-Loop task, which introduces additional visual complexity, many models could not maintain their performance, although Qwen2.5-VL-72B performed well at 31.33\%.
    Furthermore, performance dropped significantly in the most complex 2D Maze-Loop task, where even the top-scoring GPT-5 achieved only 11.04\% Acc@N.
    Although this is above the Chance Level (approx. 2.5\%), it indicates that the challenge of navigating a maze structure is extremely difficult for current VLMs.

    Additionally, the Molmo series performed at or below the Chance Level on all tasks.
    This result is strongly correlated with their notably lower Trace Coverage (Cov.).
    While other models achieved near 100\% coverage, the Molmo models remained in the 70-80\% range for many tasks.
    This is due to a high number of parse failures, suggesting the Molmo models struggle with the basic instruction-following and output formatting required by the task.

	\begin{table*}[t]
		\centering
	    \caption{Overall performance of evaluated VLMs on \ordinalbench. We report Final Accuracy (Acc@N), Stepwise Trace Accuracy (STA), Normalized Longest Correct Prefix (nLCP), and Trace Coverage (Cov.) in percentages (\%). All metrics are macro-averaged across QA instances; the highest score in each column is highlighted in \textbf{bold}. Note that results for GPT-5 and Gemini 2.5 are based on stratified subsets sampled from \texttt{public\_test} (2,500/1,500; seed = 0), while others are based on the full 15,000/9,000-sample test set.}
	        \label{tab:main_results}
	        \resizebox{\textwidth}{!}{%
        \begin{tabular}{l S[table-format=3.2] S[table-format=3.2] S[table-format=3.2] S[table-format=3.2] c S[table-format=3.2] S[table-format=3.2] S[table-format=3.2] S[table-format=3.2] c S[table-format=3.2] S[table-format=3.2] S[table-format=3.2] S[table-format=3.2]}
        \toprule
        & \multicolumn{4}{c}{\textbf{2D Single-Loop}} & & \multicolumn{4}{c}{\textbf{3D Single-Loop}} & & \multicolumn{4}{c}{\textbf{2D Maze-Loop}} \\
        \cmidrule(lr){2-5} \cmidrule(lr){7-10} \cmidrule(lr){12-15}
        \textbf{Model} & {\textbf{Acc@N}} & {\textbf{STA}} & {\textbf{nLCP}} & {\textbf{Cov.}} & & {\textbf{Acc@N}} & {\textbf{STA}} & {\textbf{nLCP}} & {\textbf{Cov.}} & & {\textbf{Acc@N}} & {\textbf{STA}} & {\textbf{nLCP}} & {\textbf{Cov.}} \\
        \midrule
        \multicolumn{15}{l}{\textit{Proprietary models}} \\
        GPT-5                  & 15.64 & 18.70 & 17.07 & 93.80 & & 13.80 & 12.05 & 8.78 & 100.00 & & \textbf{11.04} & \textbf{17.33} & \textbf{15.45} & 99.40 \\
        Gemini 2.5 Flash Lite  & 21.68 & 21.13 & 18.11 & 100.00 & & 10.93 & 9.36 & 6.75 & 100.00 & & 9.28 & 16.79 & 14.66 & 99.48 \\
        \midrule
        \multicolumn{15}{l}{\textit{Open-source models}} \\
        Qwen2.5-VL-72B         & 31.16 & 31.89 & 24.38 & 100.00 & & \textbf{31.33} & \textbf{33.25} & 22.82 & 100.00 & & 5.07 & 11.47 & 9.67 & 99.54 \\
        Qwen2.5-VL-32B         & \textbf{32.83} & 34.34 & 23.79 & 100.00 & & 30.27 & 31.93 & \textbf{23.07} & 99.93 & & 4.29 & 10.38 & 8.71 & 99.32 \\
        Qwen2.5-VL-7B          & 32.76 & \textbf{36.47} & \textbf{24.53} & 96.69 & & 14.24 & 12.10 & 9.80 & 99.80 & & 3.96 & 9.41 & 7.55 & 96.91 \\
        InternVL3.5-14B        & 32.45 & 36.38 & 22.12 & 100.00 & & 19.79 & 23.98 & 12.14 & 99.99 & & 3.23 & 9.02 & 7.28 & 99.73 \\
        InternVL3.5-8B         & 29.18 & 31.86 & 19.34 & 99.44 & & 12.36 & 13.13 & 8.70 & 99.26 & & 3.52 & 9.40 & 7.67 & 96.35 \\
        Molmo-72B              & 4.53  & 9.68  & 7.09  & 77.37 & & 6.79 & 11.33 & 6.13 & 76.80 & & 0.31 & 4.94 & 4.84 & 74.98 \\
        Molmo-7B               & 5.67  & 9.15  & 6.92  & 70.37 & & 5.41 & 8.70 & 6.14 & 89.09 & & 1.58 & 5.99 & 5.46 & 82.35 \\
        \midrule
        Chance Level & {11.70} & {$\approx 0$} & {$\approx 0$} & \multicolumn{1}{c}{---} & & {11.70} & {$\approx 0$} & {$\approx 0$} & \multicolumn{1}{c}{---} & & {2.50} & {$\approx 0$} & {$\approx 0$} & \multicolumn{1}{c}{---} \\
        \bottomrule
        \end{tabular}%
        }
    \par\smallskip{\raggedright\footnotesize\textit{Chance Level.} For the 2D/3D Single-Loop tasks we average the per-instance probability of uniform random guessing, $1/\text{\texttt{object\_count}}$, which yields 11.7\%. For the 2D Maze-Loop task we instead average $1/\text{\texttt{path\_length}}$ from each maze layout, resulting in roughly 2.5\%. STA and nLCP chance values remain near 0\% because randomly generated traces almost never match the correct step sequence.\par}
    \end{table*}

\subsubsection{Performance Analysis by Difficulty Axis}

\begin{figure*}[t]
		        \centering
		        \includegraphics[width=1.0\linewidth]{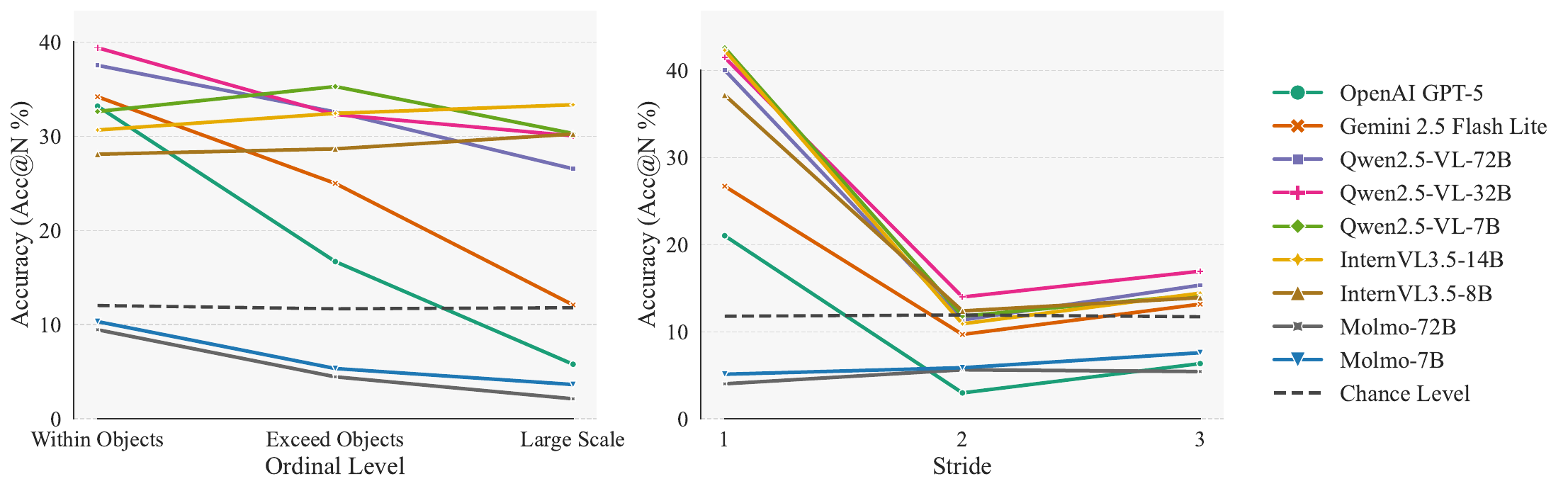}
		        \caption{
		            Performance degradation with increasing cognitive load in 2D Single-Loop.
		            (Left) Accuracy consistently declines as the ordinal level increases from Within to Large Scale.
		            (Right) Skip counting (stride $> 1$) causes a sharp drop, pushing many models toward chance level.
		        }
		        \label{fig:accuracy_vs_difficulty}
		    \end{figure*}

    \paragraph{Performance Degradation in Challenging Tasks (Maze Structures and skip counting)}
    One of the most significant findings of this study is the sharp decline in performance when models are required to perform challenging procedural reasoning.
	    Figure~\ref{fig:accuracy_vs_difficulty} (right) shows that as the cognitive load increases from simple sequential tracking (stride $= 1$) to \textbf{skip counting (stride $> 1$)}, the performance of all models drops substantially.
	    For example, in the 2D Single-Loop task, Qwen2.5-VL-7B achieved an Acc@N of 42.62\% with stride $= 1$, but this plummeted to 11.80\% with stride $= 2$, approaching the Chance Level.
    This trend is even more pronounced in the nLCP metric, which indicates the robustness of the thought process, with many models dropping to the 5--6\% range under stride $> 1$ conditions.
    This is strong evidence that models struggle to reliably execute algorithmic instructions.

\begin{figure*}[t]
		        \centering
		        \includegraphics[width=1.0\linewidth]{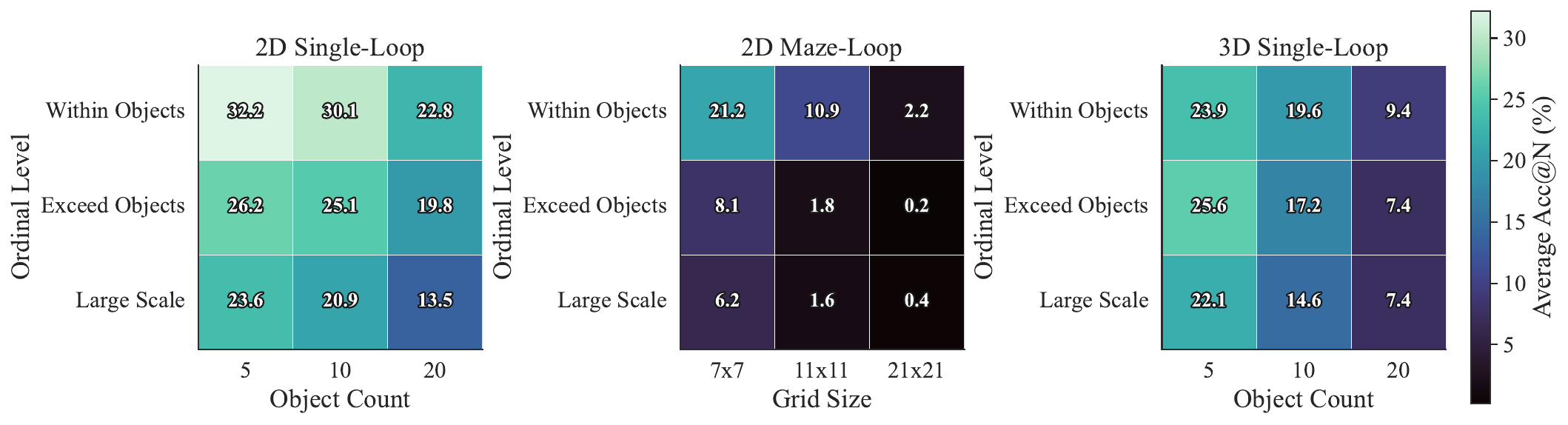}
		        \caption{
		            Average Acc@N performance heatmaps across all models in the three task categories.
		            Each heatmap displays the average accuracy of all evaluated models using color, with the y-axis representing the Ordinal Level and the x-axis representing either the number of objects or grid size.
		        }
		        \label{fig:average_accuracy_heatmaps}
		    \end{figure*}

	    \paragraph{Impact of Ordinal Magnitude}
	    Figure~\ref{fig:accuracy_vs_difficulty} (left) shows the impact of ordinal magnitude $N$ (Within / Exceed / Large Scale) on performance.
	    In the 2D Single-Loop task, many models showed a performance decline as N increased from Within Objects to Large Scale.
	    This trend becomes even more prominent in the 2D Maze-Loop task, where the average-performance heatmap in Figure~\ref{fig:average_accuracy_heatmaps} (middle) shows many models dropping to below 3\% Acc@N under Large Scale conditions.
	    This quantitatively demonstrates that VLMs have limitations in their sustained attention capacity for maintaining long-distance counts.

	    \paragraph{Impact of Scene Scale}
	    As the total number of objects or the grid size increases, there is a tendency for model performance to decrease more sharply than the corresponding drop in the Chance Level.
	    The heatmaps in Figure~\ref{fig:average_accuracy_heatmaps} show a consistent performance degradation as the number of objects (or grid size) increases.
	    Notably, in the 3D Single-Loop task, the average Acc@N drops to single digits under heavy clutter (20 objects), approaching the Chance Level (5.00\%) (Figure~\ref{fig:average_accuracy_heatmaps}, right).
	    This suggests that in large-scale scenes with increased visual clutter, it is difficult for models to maintain attention and execute an accurate counting strategy.

\section{\uppercase{Discussion}}
\label{sec:discussion}

We focus on the underlying causes of sharp performance degradation and the implications for future VLM development.

\subsection{Summary and Implications}

Our results reveal a fundamental weakness shared across all evaluated VLMs.
While many models are strong at static pattern recognition, their \textbf{procedural and algorithmic reasoning}---executing a precise, multi-step procedure sequentially over visual inputs---remains limited.
This weakness can be explained from two perspectives.

\paragraph{Language Bottleneck}
VLMs may reason by internally converting visual information into linguistic representations, which can amplify reference-resolution ambiguity \cite{zhang2025comfort}.
Complex visual structures such as mazes are hard to transcribe faithfully into language, and information loss or ambiguity during the visual-to-language conversion can cause long-horizon reasoning to break down.
Consistent failures on simple branching decisions in maze tasks suggest that models may not form stable internal representations of global spatial structure.

\paragraph{Lack of Procedural Knowledge}
Current VLMs are strong in declarative knowledge (e.g., what an object is), yet often struggle to execute \textbf{procedural knowledge (algorithms)} such as ``follow a maze traversal rule'' or ``count by skipping every three steps.''
The sharp performance drops for stride $> 1$ across all models strongly support this hypothesis and align with broader evidence of fragile generalization \cite{qi2024generalization}.
In other words, models appear to have difficulty precisely managing and updating internal state (e.g., ``which ordinal are we currently counting'') while executing procedures.

\paragraph{Implications for Robust VLM Development}
\label{sec:implications}
Based on the observed failure modes, our findings suggest that robustness may benefit from: (i) explicit, structured state representations to reduce the language bottleneck; (ii) training signals and curricula that emphasize step-level procedural execution (short to long horizons; stride $= 1$ to $> 1$); (iii) schema-constrained and validated generation to reduce parse failures; and (iv) evaluations beyond Acc@N using trace metrics (nLCP/STA/Cov.) and systematic stress tests.

Overall, for VLMs to function as ``visual agents'' rather than mere recognizers, improving the ability to accurately manipulate visual information and stably carry out multi-step reasoning is essential.

\subsection{Limitations and Future Work}

While this study provides diagnostic insights into ordinal number understanding in VLMs, it has several limitations.

\paragraph{Reliance on Synthetic Data}
\ordinalbench is synthetic by design to isolate ordinal number understanding, and thus it does not include the full complexity of real images (e.g., lighting, occlusion, and diverse textures).
Future work should extend evaluation to real-world images where recognition and reasoning are intertwined.

\paragraph{Static Environments}
Our evaluation targets still images and does not address dynamic situations that involve temporal ordering (e.g., videos).
Reasoning evaluation in dynamic environments is an important direction for applications such as autonomous driving and robotics.

\section{\uppercase{Conclusion}}
\label{sec:conclusion}
In this work, we present \ordinalbench, a large-scale diagnostic benchmark for evaluating VLMs' generalization in ordinal number understanding.
It provides systematic difficulty control across 2D Single-Loop, 2D Maze-Loop, and 3D Single-Loop tasks, a skip counting task that probes procedural reasoning, and an evaluation protocol with nLCP and STA that scores step-by-step traces beyond final accuracy.

Across extensive experiments, even state-of-the-art VLMs show marked degradation under complex path tracing and algorithmic conditions, revealing a core limitation.
We expect \ordinalbench to serve as a reliable yardstick that catalyzes progress toward more robust, instruction-following VLMs and, ultimately, more general visual intelligence.

\section*{\uppercase{Acknowledgements}}
\label{sec:acknowledgements}
This work was partially supported by JSPS KAKENHI Grant Number 23K11342.

We used ChatGPT (OpenAI) to assist with drafting parts of this manuscript and polishing English expressions; the authors take full responsibility for the final content.

\end{document}